\documentclass[a4paper]{article}

\usepackage[english]{babel}
\usepackage[utf8]{inputenc}
\usepackage{amsmath}
\usepackage[margin=1.5in]{geometry}
\usepackage{graphicx}
\usepackage[colorinlistoftodos]{todonotes}
\usepackage{indentfirst}

\title{Detecting Deepfake Videos:\\ An Analysis of Three Techniques}

\author{
    Armaan Pishori\\
    \texttt{pishori@usc.edu}
    \and 
    Brittany Rollins\\
    \texttt{bmrollin@usc.edu}
    \and 
    Nicolas van Houten\\
    \texttt{nvanhout@usc.edu}
    \and 
    Nisha Chatwani\\
    \texttt{nchatwan@usc.edu}
    \and 
    Omar Uraimov\\
    \texttt{uraimov@usc.edu}
}

\date{\today}

\begin{document}
\maketitle

\begin{abstract}
Recent advances in deepfake generating algorithms that produce manipulated media have had dangerous implications in privacy, security and mass communication. Efforts to combat this issue have risen in the form of competitions and funding for research to detect deepfakes. This paper presents three techniques and algorithms- convolutional LSTM, eye blink detection and grayscale histograms- pursued while participating in the Deepfake Detection Challenge. We assessed the current knowledge about deepfake videos, a more severe version of manipulated media, and previous methods used, and found relevance in the grayscale histogram technique over others. We discussed the implications of each method developed and provided further steps to improve the given findings.
\end{abstract}

\section{Introduction}

Deepfakes are videos or images that have been manipulated to appear different from their original state. Although videos and images have been altered with technologies such as Adobe Premiere Pro, Final Cut Pro, and others for decades, deepfakes differ from these altered media because they are created using machine learning. The “deep” in deepfake is indicative of the subcategory of machine learning under which deepfakes reside: deep learning.

\begin{figure}[ht]
\centering
\includegraphics[width=0.7\textwidth]{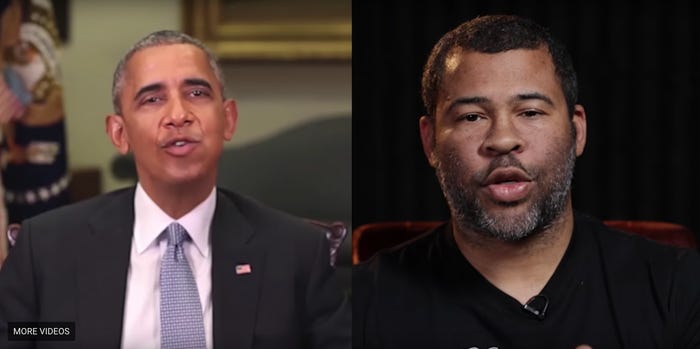}
\caption{Deepfake of President Barack Obama made with Lip-Sync Technique \cite{Obama}}
\label{fig:Obama}
\end{figure}

\par
The effects of deepfake technology range from harmless to dangerous depending on the user’s intent. Notably, a video went viral in which President Barack Obama appears to be insulting public figures, including Donald Trump and Ben Carson \cite{Obama}. As shown in Figure \ref{fig:Obama}, the video was made by manipulating President Obama's lips to match Jordan Peele's voice. Although this video was created to raise awareness of the dangers of this technology, deepfake videos with malicious intent and without warnings still do exist. Such videos may leave viewers with negative sentiments about the speaker being misrepresented. Misuse of this technology contributes to a dangerous spread of misinformation \cite{danger}.

\par
To combat the spread of misinformation, industry leaders and researchers have fostered a growing interest in developing computational approaches for detecting deepfakes. In January 2020, our research team participated in the Kaggle Deepfake Detection Competition \cite{Comp}. The aim of the competition was to improve deepfake detection technology by releasing large amounts of new data in a competitive setting. Participants were provided with 500 GB of video data that consisted of both real and fake videos. The competition scored competitors based on their log loss, which measures the number of false classifications made by the classifier. If a log loss is closer to zero, then the classifier is more accurate. The winner of the challenge achieved a log loss of 0.19207 on the public leaderboard \cite{CompRes}. This result proves the difficulty of exposing deepfakes because there is still room for error in identifying them. Our research focused on implementing different preprocessing approaches and analyzing their effectiveness in detecting deepfakes. Our group utilized several different techniques —implementing a basic convolutional neural network and recurrent neural network architecture (CNN+RNN), detecting the number of eye blinks, and preprocessing the data using grayscale histograms— in an attempt to create a better detector and mitigate the spread of misinformation. This paper demonstrates the following findings: using a CNN+RNN model provides great results due to artifact identification, and there is potential for grayscale histograms to give high accuracy with proper training and time. The paper is organized as follows: Section 2 contains related works that contributed to this research, Section 3 describes our approach to preprocessing the data, Section 4 illustrates our CNN + RNN architecture, Section 5 describes the use of eye blink data and detection to inform models, Section 6 contains the grayscale histograms approach, Section 7 compares the results of the different approaches pursued, Section 8 details how the group could further the research described in the paper, and Section 9 contains the conclusion.

\section{Related Work}

\subsection{Creating Deepfake Videos} 

\begin{figure}[ht]
\centering
\includegraphics[width=0.7\textwidth]{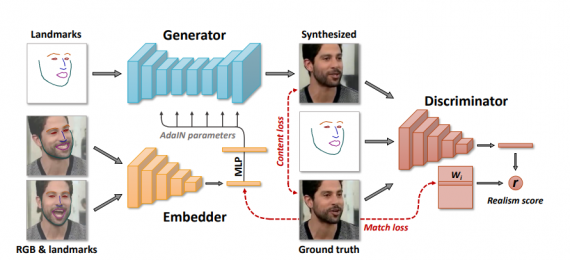}
\caption{A diagram of GAN architecture \cite{GAN-arch}}
\label{fig:GAN}
\end{figure}

Deepfake videos are created using Generative Adversarial Networks (GAN), which are a subset of deep learning. To create videos, GANs are fed input video data, from which they can generate a unique video. The adversarial part of the network comes from a discriminative network that checks whether the generated video seems authentic. The discriminative network and the generator work against one another to improve the generated videos until the generator fools discriminator. Once this is achieved, the generator presumably created a deepfake that can fool humans \cite{GAN}. Within the last few years, deepfakes have grown in popularity and accessibility. Notably, an application called FakeApp gives people the ability to easily create deepfakes \cite{FaceApp}. By using thousands of videos and images of a desired person’s face, this app can generate convincing deepfakes. As of July 2019, the FaceApp has received over 120 million downloads, proving the popularity of this technology \cite{FaceAppStat}. Since deepfake technology has become accessible, the number of generated videos have grown enormously.

\subsection{Convolutional LSTM} 

\begin{figure}[ht]
\centering
\includegraphics[width=0.7\textwidth]{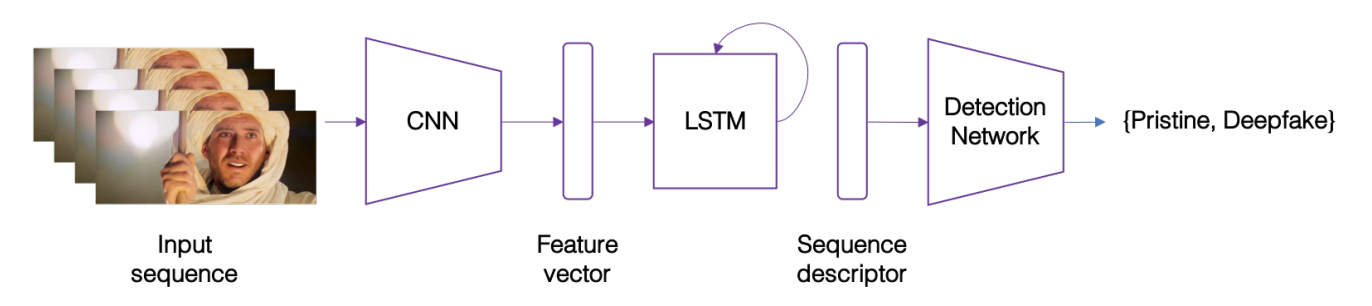}
\caption{Overview of LSTM architecture \cite{Delp}}
\label{fig:lstm}
\end{figure}

Past deepfake detection models utilized a convolutional LSTM structure which combines a CNN for “frame feature extraction” and LSTM for “temporal sequence analysis” proposed by Delp et al \cite{Delp}. With this model, they were able to divide the tasks for detecting deepfake videos by specifically focusing on image manipulation in each frame of the video. With a total dataset of 600 videos, Delp et al. performed a random split allocating 70 percent of the video for training, 15 percent for validation, and 15 percent for testing. After resizing the frames and reducing the input to only the frames that are necessary for detection in each video, they were able to determine if a video was a deepfake or not with an accuracy above 97 percent with only 40 frames from each video \cite{Delp}.

\subsection{Eye Blinking}

Eye Blinking is known as a fundamental biological function that is extremely hard to emulate in deepfake videos. With an average rate of 4.5 blinks per second and each blink lasting 0.1-0.4 seconds, most training datasets of videos used for deepfake detection have a scarce amount of faces with their eyes closed. Therefore, the lack of eye blinking can be a promising indicator of a deepfake video. Chang et al. explored this concept of exposing deepfake videos with eye blink detection by using a Long-term Recurrent Convolutional Network (LCRN) to integrate the temporal relationship between video frames from the time the eye opens to when it closes \cite{eye}. Their preprocessing involved locating facial landmarks using a face detector and removing the background around the eyes. In their experiment, they were able to obtain a 99\% accuracy with detecting deepfake videos using their LCRN which was higher than the performance of their CNN. This is due to the fact that their CNN was unable to incorporate the temporal knowledge of previous frames and their LCRN could memorize temporal knowledge of past states and predict if the eyes would be open or closed in the next few frames of the video.

\subsection{Spectral Responses}

\begin{figure}[ht]
\centering
\includegraphics[width=0.7\textwidth]{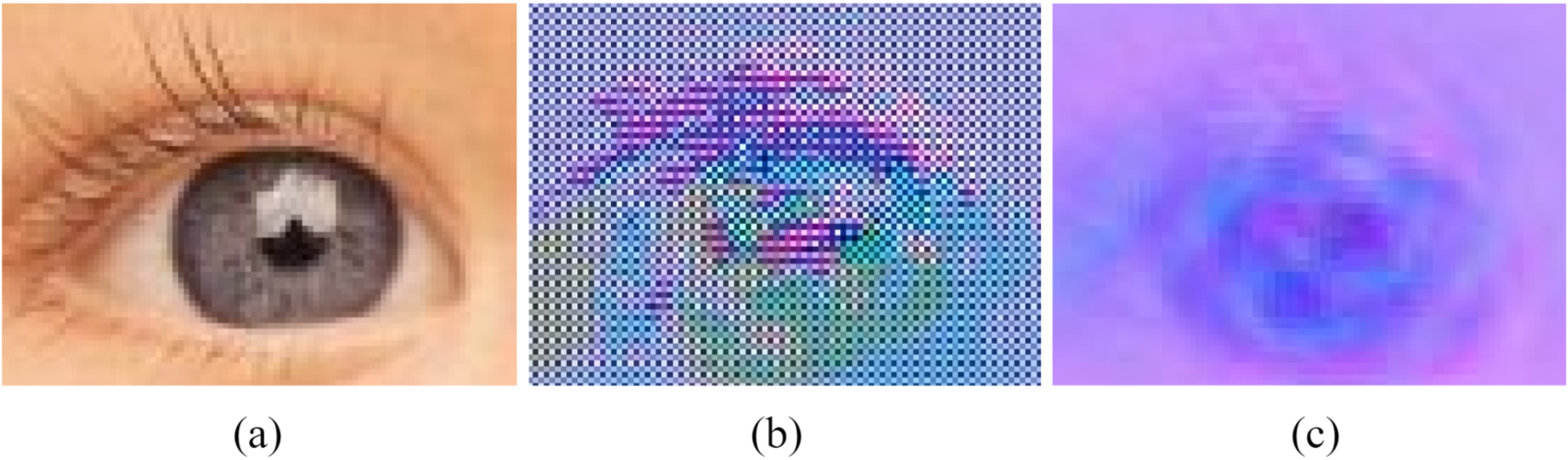}
\caption{Demonstration of how advancements in deconvolution steps have reduced checkerboard artifacts \cite{deconv_img}. (a) shows the original image (b) shows the results for a sub-pixel convolution (c) shows the results for a more complex repeat sub kernel initialized sub-pixel convolution}
\label{fig:Checkerboard_Eyes}
\end{figure}

Deepfake preprocessing research is focused on highlighting artifacts left behind by deconvolution steps in GAN-generated imagery. For instance, a commonly sought out artifact is the checkerboard pattern that is left behind while rebuilding GAN-generated imagery \cite{deconv_img}\cite{Color_Cues}\cite{Checker_Board}. Artifacts like these have been monumental in the fight to detect deepfakes, however, detection models will not be able to rely on them forever. Since artifacts like these are leftover from the deconvolutional process instead of variances in the images, it is relatively easy to isolate and fix them. For example, researchers have already been able to mitigate the checkerboard artifact by incorporating up-sampling followed by convolution steps into their model \cite{Checker_Board}. Therefore, there is more long term value in highlighting complex artifacts that are more strongly tied to the inputted video instead of the GAN model being used. These preprocessing methods make it more difficult for adversaries to fix the issue by adding a simple update to their model structure.

\begin{figure}[ht]
\centering
\includegraphics[width=0.7\textwidth]{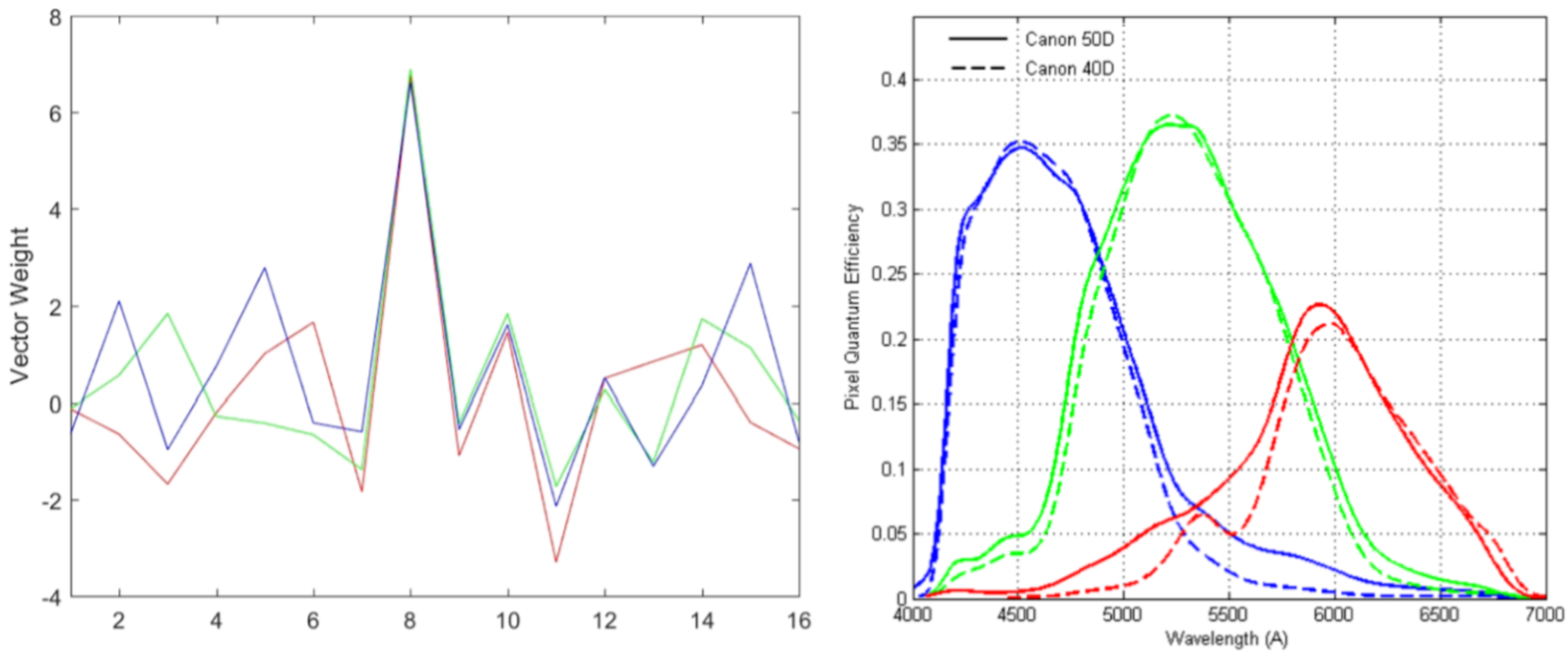}
\caption{(Left) The weights learned by a GAN face synthesis model \cite{GAN_weights} (Right) The spectral response of two different Canon cameras \cite{Canon}}
\label{fig:Spectral_Responce}
\end{figure}

Extracting color histograms out of images and videos is a preprocessing method that is difficult to counteract. This is because color histograms are not only unique to their derived images, but they are also unique to the camera the image was taken on—both of which are independent to the deepfake model being used \cite{Color_Cues}. The spectral response of a camera’s filter varies by the camera model \cite{Canon}. Therefore, GAN-generated components in images do not only have to mimic the current color composition of the image, but they must also mimic the underlying spectral response of the camera. discrepancy between GAN-generated spectral response graphs and actual camera spectral response graphs vary by a significant amount as seen in Figure \ref{fig:Spectral_Responce}. This discrepancy can be measured by generating a color histogram sequence of a deepfake video.

\par
Similarly, a grayscale histogram can be used in replacement of a color histogram in order to alleviate the computational stress of having a color dimension. Additionally, this method has yielded superior detection results in past research on GAN-generated imagery conducted by McCloskey et al.\cite{Color_Cues}. They speculated that their color histogram model was too sensitive and therefore flagged images with common Photoshop touch-ups as GAN-generated imagery. Therefore, reducing the data dimension by using a grayscale histogram may have made the model more robust against false positives.

\section{Preprocessing Methodology}

Our research focused on assessing the effectiveness of different preprocessing approaches for deepfake detection models, and this section describes our methodology for doing so. Additionally, There are an endless number of factors that could be taken into consideration when creating a preprocessing system. Therefore, Our team decided to narrow our focus to ensuring that our system could utilize the entire 500 GB data set under reasonable computing restraints. With this in mind, we tested multiple preprocessing methods, some to failure, before committing to them.

\par
Ultimately, our research narrowed down to two different preprocessing techniques alongside an CNN+RNN model that did not utilize any special preprocessing and acted as a control method. The first preprocessing method highlighted differences between a camera’s unique fingerprint and a GAN’s fingerprint. The second method utilized biosignals, specifically eye blinks, as a more complex indicator for deepfake videos. The results of these two techniques were then compared to our control method in order to see whether the extra computational resources required for preprocessing were worth it.

\section{Method 1: CNN + RNN}

\begin{figure}[ht]
\centering
\includegraphics[width=0.7\textwidth]{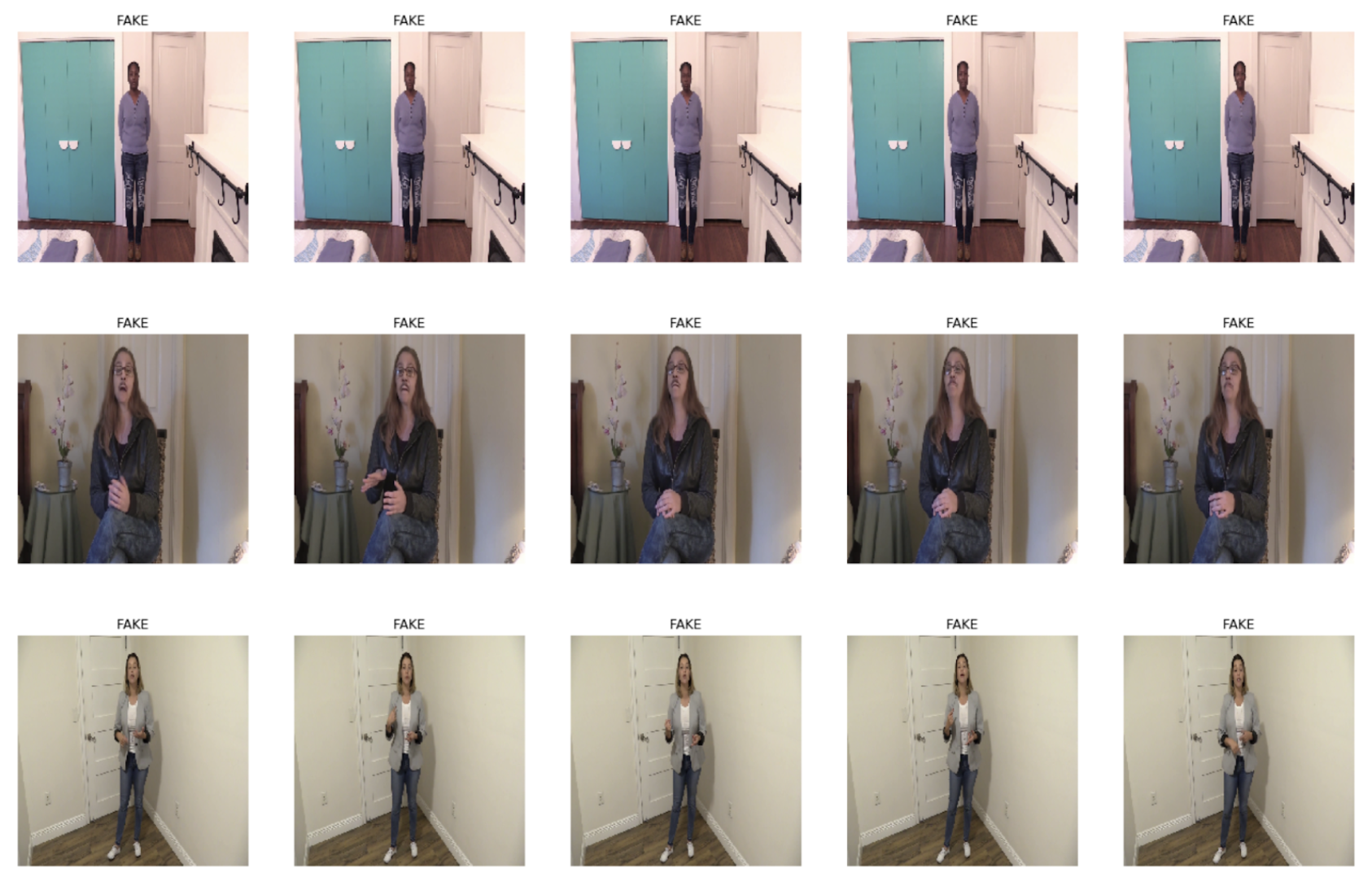}
\caption{An example output of the video frame generator using the data set}
\label{fig:Frame Generator}
\end{figure}

To begin our model building process, it was necessary to investigate how to build a basic convolutional LSTM model, as the data we were provided came in short video clip form and would require both temporal and spacial frame analysis. For this model, no preprocessing techniques were pursued, outside of finding a way to input the video data into the model.

\par
The first issue occurred when the model was trying to find a way to input the video data. This obstacle was overcome by using Ferlet's video frame generator that would take video inputs and transform them into a sequence of image frames that could then be fed into our CNN+RNN sequence \cite{Frame Generator}. The model inputted a sequence of images of size 256 by 256 pixels through 12 layers of a CNN before passing the sequence through an additional 4 layers of an LSTM. This basic model was run on the raw video data without any preprocessing.

\section{Method 2: Eye Blink}

\subsection{Eye Blink Statistics}

After experimenting with initial models, the next model specifically focuses on one feature of the face that could be beneficial in detecting deepfake videos. There are very few images online of faces with their eyes closed. As a result, it is more difficult for programmers to create deepfake videos that accurately emulate the human rate of blinking. Table \ref{table:1} demonstrates the discrepancies of the average number of blinks between actual people, the people in the real videos, and the people in the fake videos. It is clear that the average number of blinks in 10 seconds is much less for fake videos, so the model was able to capitalize on this feature.

\begin{table}[ht]
\centering
\begin{tabular}{ |c|c|c| } 
\hline
Real Life & Real Videos & Fake Videos \\
\hline
 3.4 & 4.8 & 2.2 \\ 
 \hline
\end{tabular}
\caption{Average number of blinks for people in real life, the given real videos, and the given deepfake videos}
\label{table:1}
\end{table}

\subsection{Architecture}

\begin{figure}[ht]
\centering
\includegraphics[width=0.7\textwidth]{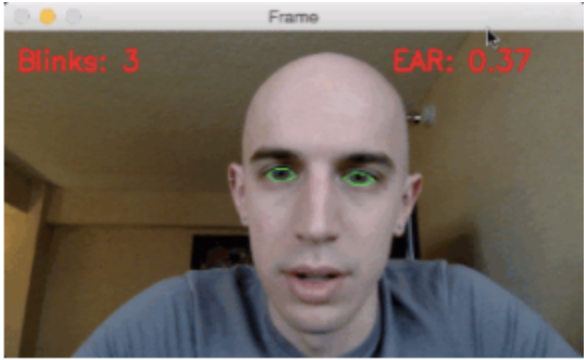}
\caption{An example of a video frame with the eye blink counter \cite{opencv}}
\label{fig:blink_frame}
\end{figure}

A classification model that relies on the number of eye blinks in the video was created. The model used OpenCV \cite{opencv} to detect facial landmarks and count the number of eye blinks as shown in Figure \ref{fig:blink_frame}, and the Eye Aspect Ratio (EAR) was recorded as well to determine the time lapse between each blink. The script to count the number of eye blinks was recorded with the training set of videos, and the model used this data to make predictions about the test set using a KNN classifier in order to determine if a video was a deepfake or not. 

\section{Method 3: Grayscale Histograms}

\subsection{Preprocessing}

\begin{figure}[ht]
\centering
\includegraphics[width=0.7\textwidth]{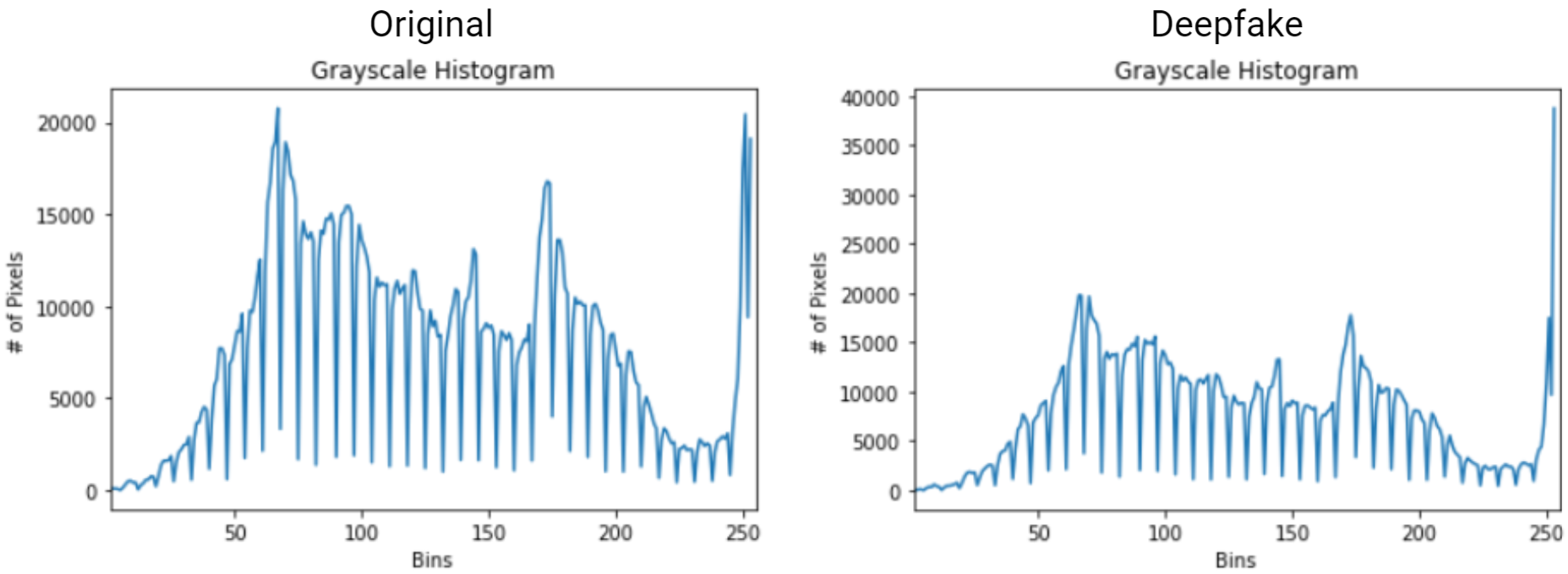}
\caption{(Left) Grayscale histogram of an unaltered video (Right) Grayscale histogram of a face swap deepfake of the same video}
\label{fig:Grayscale_Histograms}
\end{figure}

Previous research conducted by McCloskey et al.\cite{Color_Cues} has shown that there is a learnable difference between GAN-generated spectral response graphs and camera-generated spectral response graphs. Additionally, when testing multiple preprocessing methods they found grayscale histograms to be the most effective at highlighting spectral response differences. Within the scope of our dataset, a clear difference can be seen when testing this preprocessing method on a video and its deepfake counterpart from our training set as seen in Figure \ref{fig:Grayscale_Histograms}. Taking this into consideration, our third model’s preprocessing system attempts to extend upon the findings of McCloskey et al.  \cite{Color_Cues}. by extracting a series of grayscale histograms out of each video. Extracting a series of grayscale histograms will allow our model to not only analyze each frame’s spectral response but also analyze how the spectral responses change over time. These series of histograms are then normalized to make sure that all histograms have an equal influence on the model.

\subsection{Architecture}

Our model sought to improve on the model created by McCloskey et al. \cite{Color_Cues} by expanding the input space of our neural network to include a temporal dimension. This was achieved by implementing a 64 neuron LSTM layer into our model. This addition enabled our model to break up each inputted video’s 300 grayscale histograms into smaller batches of 10 histograms while maintaining the temporal relationship from the original, larger sequence. The LSTM layer then outputted its results into two more neural network layers that would increment then decrement in size from 128 neurons to 64 neurons, thus ultimately resulting in a final classification for the video. When experimenting with how many layers optimized our model’s accuracy, it was found that a more simplistic neural network helped improve the accuracy. This was a reasonable finding considering how histograms are not a very complex data type.

\section{Results}

Each of the three approaches discussed was tested using different model architectures and parameter optimizations in order to maximize their accuracy. Furthermore, each approach outputted a standardized result to help facilitate comparisons. In the table below, these results can be seen alongside the RNN+CNN model, which acted as a control model since it did not utilize any preprocessing. This comparison demonstrates the impact of investing resources into a preprocessing system.

\begin{table}[ht]
\centering
\begin{tabular}{ |c|c|c|c| } 
\hline
Model & Accuracy & Val Loss & Val Accuracy \\
\hline
CNN+RNN & 82.20\% & 1.6847 & 82.81\% \\ 
\hline
Eye Blink Detection & 81.67\% & 0.4762 & 81.67\% \\ 
\hline
Grayscale Histogram & 85.71\% & 0.5927 & 81.32\% \\ 
\hline
\end{tabular}
\caption{Displays the final accuracy, validation loss and accuracy for each model}
\label{table:2}
\end{table}

From Table \ref{table:2}, it can be inferred that the grayscale histogram had the highest accuracy and lowest validation loss compared to others, however, all models ranged within the 80-90\% range with slight improvements from the control model. Each one had its own limitations which might have restricted the final outcome. Due to computational limitations of the resources at hand, about 50GB of the total dataset was used to train most models which contributed to lower accuracies than expected. The CNN+RNN model had performed the best with the base unaltered subset of the dataset, as it identified artifacts left by deepfake GANs in areas of the frame even without the character's face. As for the eye blink detection model, an analysis of the average eye blink dataset generated from our video dataset showed that a few outliers amongst the faked videos could have been removed before training the model. Generating the grayscale histograms for each video was a computational stress that limited the capacity of videos more so than the other models. However, the results being on par with other models suggests that the grayscale histogram model can perform with a much higher accuracy given a larger dataset.

\section{Discussion and Further Research}

Our findings indicate that preprocessing system are key components of high performing deepfake detection models. For instance, in the discovery of irregular eye blinks being a universal deepfake indicator, Chang et al. was able to achieve a 99\% detection accuracy at the time \cite{eye}. Unfortunately, our findings suggest that deepfake generators are improving rapidly and a variety of detection methods, such as eye blink detection, have lost their effectiveness. Fortunately, there are many more complex biosignals that have yet to be rigorously tested. For instance, Demir et al. discuss how the medical community has been researching robust biosignals long before deepfakes were invented. Complex biosignals such as tracking subtle changes of color and motion in cheeks to infer heart rates may prove to be just as effective as eye-blink tracking once was \cite{biosignals}. 

\par
In addition to pursuing new indicators, research should also be invested in indicators that have already shown potential for growth, such as spectral response variations that can be seen through grayscale histograms. Adding a temporal dimension to our grayscale histogram model significantly improved the accuracy compared to previous grayscale histogram-based detectors. For instance, when using grayscale histograms to detect fake images, McCloskey et al. were able to achieve a 61\% accuracy \cite{Color_Cues}. Even though our model in comparison to previous methods yielded promising results, there is still room for improvement. By having our model input grayscale histograms it loses all spatial awareness. Considering that deepfakes are generally concentrated in one region of each video frame, our model is unable to focus solely on this region. Instead, it is forced to analyze the frame as a whole, thus mixing data regions of higher importance and lower importance. Therefore, extracting grayscale histograms from augmented regions such as faces instead of the entire frame has the potential to significantly improve our accuracy.

\section{Conclusion}

In conclusion, our results indicate an advent of newer techniques as the grayscale histogram and the ability to build more robust models from existing models to combat deepfake media. The limitations stated above can be mitigated with access to more resources, which would lead to better results. These methods can further be combined to develop more accurate model and bring us closer to combating this issue. Understanding the implications of spreading misinformation is crucial in solving the problem at hand. Many aspects of our society run on the validity of information especially life-altering decisions, such as forensic and legal video analysis. Trust is a pertinent factor in living in a collaborative society. By losing this trust in media, we stray further away from it leading to public's misrepresentation of emotions. Disregarding the backlash, Sp.a was able to showcase this loss of trust by generating a deepfake of President Trump spreading controversial advice on the issue of climate change \cite{Misinformation}. Taking the appropriate immediate steps to diminish deepfake media's effectiveness can prevent it's serious ramifications in the future.

\section*{Acknowledgement}

We thank the USC Center for Artificial Intelligence in Society (CAIS) and the Center for Artificial Intelligence in Society's Student Branch (CAIS++) for providing insight, expertise and resources which assisted in our research and development of our findings.

\par
We would also like to thank Patrick Darrow for providing feedback throughout this process which helped improve our models and manuscript.


\newpage

\begin{thebibliography}{9}

\bibitem{deconv_img}
    Aitken, Andrew, et al. “Checkerboard Artifact Free Sub-Pixel Convolution: A Note on Sub-Pixel Convolution, Resize Convolution and Convolution Resize.” ArXiv:1707.02937 [Cs], July 2017. arXiv.org, http://arxiv.org/abs/1707.02937.

\bibitem{biosignals}
    Aybars Ciftci, Umur, et al. “FakeCatcher: Detection of Synthetic Portrait Videos Using Biological Signals.” ArXiv.org, 9 Aug. 2019, arxiv.org/pdf/1901.02212.pdf.

\bibitem{Canon}
    Buil, Christian. “Canon 40D, 50D, 5D, 5D Mark II Comparison.” Astrosurf, 2009, astrosurf.com/buil/50d/test.htm.
    
\bibitem{COCO}
    Lin, Tsung-Yi, et al. “Microsoft COCO: Common Objects in Context.” ArXiv:1405.0312 [Cs], Feb. 2015. arXiv.org, http://arxiv.org/abs/1405.0312.

\bibitem{Obama}
    BuzzFeedVideo. (2018, April 17). You Won’t Believe What Obama Says In This Video! [Video]. Youtube. https://www.youtube.com/watch?v=cQ54GDm1eL0 

\bibitem{Comp}
    “Deepfake Detection Challenge.” Kaggle, https://www.kaggle.com/c/deepfake-detection-challenge/overview.

\bibitem{Delp}
    D. Güera and E. J. Delp, "Deepfake Video Detection Using Recurrent Neural Networks," 2018 15th IEEE International Conference on Advanced Video and Signal Based Surveillance (AVSS), Auckland, New Zealand, 2018, pp. 1-6, doi: 10.1109/AVSS.2018.8639163.

\bibitem{FaceApp}
    FaceApp. FaceApp Inc, 2016. Vers 4.013. Apple App Store, https://apps.apple.com/gh/app/faceapp-ai-face-editor/id1180884341

\bibitem{Frame Generator}
    Ferlet, Patrice. Training neural network with image sequence, an example with video as input. 26 Nov. 2019. medium.com,  https://medium.com/smileinnovation/training-neural-network-with-image-sequence-an-example-with-video-as-input-c3407f7a0b0f

\bibitem{danger}
    Galston, William A. “Is Seeing Still Believing? The Deepfake Challenge to Truth in Politics.” Brookings, Brookings, 6 May 2020, www.brookings.edu/research/is-seeing-still-believing-the-deepfake-challenge-to-truth-in-politics/.

\bibitem{GAN}
    Goodfellow, Ian J. Jean Pouget-Abadie, Mehdi Mirza, Bing Xu, David Warde-Farley,
    Sherjil Ozair, Aaron Courville and Yoshua Bengio, "Generative Adversarial Nets," Advances in Neural Information Processing Systems 27 (NIPS 2014), Montreal, Canada, 2014, pp. 1-9, https://arxiv.org/pdf/1406.2661.pdf

\bibitem{FaceAppStat}
    Iqbal, Mansoor. “FaceApp Revenue and Usage Statistics (2020).” Business of Apps, 23 June 2020, https://www.businessofapps.com/data/faceapp-statistics.

\bibitem{GAN_weights}
    Karras, Tero, et al. “Progressive Growing of GANs for Improved Quality, Stability, and Variation.” ArXiv:1710.10196 [Cs, Stat], Feb. 2018. arXiv.org, http://arxiv.org/abs/1710.10196.

\bibitem{Color_Cues}
    McCloskey, Scott, and Michael Albright. “Detecting GAN-Generated Imagery Using Color Cues.” ArXiv:1812.08247 [Cs], Dec. 2018. arXiv.org, http://arxiv.org/abs/1812.08247.

\bibitem{Checker_Board}
    Odena, Augustus, et al. “Deconvolution and Checkerboard Artifacts.” Distill, 17 Oct. 2016, https://distill.pub/2016/deconv-checkerboard.

\bibitem{CompRes}
    Rosebrock, Adrian.“Deepfake Detection Challenge Results.” Kaggle, https://www.kaggle.com/c/deepfake-detection-challenge/leaderboard
    
\bibitem{opencv}
   Rosebrock, Adrian. “Eye Blink Detection with OpenCV, Python, and Dlib.” PyImageSearch, 24 Apr. 2017, https://www.pyimagesearch.com/2017/04/24/eye-blink-detection-opencv-python-dlib.
    
\bibitem{Misinformation}
    Schwartz, Oscar. “You Thought Fake News Was Bad? Deep Fakes Are Where Truth Goes to Die.” The Guardian, Guardian News and Media, 12 Nov. 2018, https://www.theguardian.com/technology/2018/nov/12/deep-fakes-fake-news-truth.

\bibitem{eye}
    Y. Li, M. Chang and S. Lyu, "In Ictu Oculi: Exposing AI Created Fake Videos by Detecting Eye Blinking," 2018 IEEE International Workshop on Information Forensics and Security (WIFS), Hong Kong, Hong Kong, 2018, pp. 1-7, doi: 10.1109/WIFS.2018.8630787.

\bibitem{GAN-arch}
    Zakharov, Egor. 2019. Digital Image. \textit{Few-Shot Adversarial Learning of Realistic Neural Talking Head Models}. 15 June 2020. https://arxiv.org/abs/1905.08233

\end{thebibliography}
\end{document}